\pdfoutput=1

\documentclass[11pt]{article}

\usepackage{acl}

\usepackage{times}
\usepackage{latexsym}
\usepackage{xcolor}
\usepackage{graphicx}
\usepackage{pifont}
\usepackage{adjustbox}
\usepackage{booktabs}
\usepackage{multicol}
\usepackage{multirow}
\usepackage{floatrow}

\newcommand{\checkmark}{{\color{teal} \ding{52}}}
\newcommand{\xmark}{{\color{red} \ding{55}}}
\newcommand{\qmark}{{\fontfamily{cyklop}\selectfont {\color{brown} \textit{?}}}}

\usepackage{algorithm}
\usepackage{algpseudocode}

\usepackage{amsmath}

\usepackage[T1]{fontenc}

\usepackage[utf8]{inputenc}

\usepackage{microtype}

\title{How Hard is this Test Set? \\NLI Characterization by Exploiting Training Dynamics}

\author{Adrian Cosma$^1$, Stefan Ruseti$^1$, Mihai Dascalu$^1$, Cornelia Caragea$^2$ \\
    {$^{1}$National University of Science and Technology POLITEHNICA Bucharest, București, Romania} \\
    {$^2$University of Illinois at Chicago, Chicago, IL} \\
    {\small\texttt{\{ioan\_adrian.cosma, stefan\_ruseti, mihai\_dascalu\}@upb.ro}}\\
    {\small\texttt{cornelia@uic.edu}}\\}

\begin{document}
\maketitle
\begin{abstract}
Natural Language Inference (NLI) evaluation is crucial for assessing language understanding models; however, popular datasets suffer from systematic spurious correlations that artificially inflate actual model performance. To address this, we propose a method for the automated creation of a challenging test set without relying on the manual construction of artificial and unrealistic examples. We categorize the test set of popular NLI datasets into three difficulty levels by leveraging methods that exploit training dynamics. This categorization significantly reduces spurious correlation measures, with examples labeled as having the highest difficulty showing markedly decreased performance and encompassing more realistic and diverse linguistic phenomena. When our characterization method is applied to the training set, models trained with only a fraction of the data achieve comparable performance to those trained on the full dataset, surpassing other dataset characterization techniques. Our research addresses limitations in NLI dataset construction, providing a more authentic evaluation of model performance with implications for diverse NLU applications.
\end{abstract}

\section{Introduction}
\label{sec:intro}

Natural Language Inference (NLI), or textual entailment \cite{dagan2009}, has emerged as an enduring challenge in the field of Natural Language Processing for evaluating the Natural Language Understanding (NLU) capabilities of models. Persistently, NLI remains a difficult problem, as it implies reasoning across several linguistic phenomena to determine the logical relationship (i.e., entailment, contradiction, or neutral) between two documents - a premise and a hypothesis. The capability to accurately infer relationships between sentences is crucial for a wide range of applications, such as question answering \cite{qanli}, dialogue systems \cite{welleck-etal-2019-dialogue}, and fact-checking \cite{thorne-etal-2018-fever,stab2018cross}. 

Since its inception \cite{nli-first}, several large-scale benchmark datasets have been proposed for NLI \cite{snli:emnlp2015,multi-nli,nie2019adversarial}; however, in time, the most ubiquitously used are the Stanford Natural Language Inference (SNLI) \cite{snli:emnlp2015} and the MultiNLI datasets \cite{multi-nli}, which have played a pivotal role in advancing the state of the art \cite{nlisurvey2019}.

However, multiple works \cite{liu2020hyponli,gururangan-etal-2018-annotation,tsuchiya-2018-performance,poliak-etal-2018-hypothesis,naik-etal-2018-stress,glockner-etal-2018-breaking} pointed out several critical limitations in these datasets, stemming from systematic annotation errors and spurious correlations that impact both the training and test sets. A critical consequence of these issues is the inflation of model performance, leading to seemingly high results \cite{naik-etal-2018-stress,liu2020hyponli} that may not generalize well to real-world scenarios. For example, a widely used RoBERTa model \cite{liu2019roberta} trained \textit{solely on the hypothesis} achieves an unreasonable accuracy of 71.7\% on SNLI and 61.4\% on MultiNLI (random chance being 33\%), which strongly points towards systematic errors in dataset construction.

In this work, we aim to address the limitations of existing NLI datasets by proposing an automated construction of a more challenging test set. In contrast to previous approaches, we avoid manually creating artificial examples \cite{naik-etal-2018-stress}; instead, we leverage existing samples from the test set. To accomplish this, we generalize dataset cartography \cite{swayamdipta2020dataset} to cluster samples in the test set and characterize them into three categories of increasing difficulty. Our approach leverages 8 measures of training dynamics of each premise-hypothesis pair and is inspired by related works in both NLI \cite{naik-etal-2018-stress,geiger2018stress,liu2020hyponli} and approaches tackling the problem of learning with noisy data \cite{pleiss2020identifying,swayamdipta2020dataset}. We show that our method can isolate examples exhibiting spurious correlations and provide a challenging test set. Furthermore, our method is general, model-agnostic, and easily extensible to other datasets (e.g., for fact-checking \cite{thorne-etal-2018-fever}). Our experiments show that using the same method on the training set enables the aggressive filtering of uninformative examples during training, reducing data quantity but increasing quality, enabling the model to obtain on-par performance on the NLI stress test proposed by \citet{naik-etal-2018-stress}, using only a fraction of data. We make our code publicly available\footnote{\href{https://github.com/cosmaadrian/nli-stress-test}{https://github.com/cosmaadrian/nli-stress-test}}.

This work makes the following contributions:
\begin{enumerate}
    \item We denote spurious correlations in the test sets for two popular NLI datasets - SNLI \cite{snli:emnlp2015} and MultiNLI \cite{multi-nli} and a fact-checking dataset, repurposed for NLI: FEVER \cite{thorne-etal-2018-fever}. We show statistically significant correlations between the performance of models and the presence of several measures of spurious correlations across labels.

    \item We propose a general method for creating a strong test set for NLI. Using a multitude of training dynamics features of samples in an existing test set, our method automatically characterizes examples in the test set into three increasing difficulty levels, which strongly correlate with decreased model performance. Our method minimizes spurious correlations, providing a more accurate measure of model performance in the real world on NLU tasks. Our method is model-independent and the underlying difficulty splits generalize across models.

    \item The same procedure applied to the training data achieves similar performance on the test set while using only 33\% of the available data for SNLI and 59\% for MultiNLI, surpassing other dataset characterization methods \cite{pleiss2020identifying,swayamdipta2020dataset}, indicating that our approach can be used as a strong method for increasing data quality.
\end{enumerate}

The paper is structured as follows. After emphasizing the shortcomings of existing NLI datasets and presenting various stress tests, we introduce our method for test set characterization. Then, we present the main results, a comparison with a different encoder to argue that our approach is model-agnostic, and an analysis supporting the viability of our approach as an alternative to training set characterization. The paper ends with conclusions and limitations.

\section{Related Work}
\label{sec:related}
Across the development of natural language inference and understanding systems, multiple large-scale training and testing datasets have been developed over different linguistic domains. Initially, progress was driven by the addition of SNLI \cite{snli:emnlp2015}, but several other variants have been proposed, such as MultiNLI \cite{multi-nli}, containing multiple domains, SciNLI \cite{sadat-caragea-2022-scinli} for scientific question answering, SQuAD \cite{rajpurkar-etal-2016-squad} and GLUE \cite{wang-etal-2018-glue} benchmarks for general-purpose NLU. Moreover, many related problems in NLU can be cast as an NLI problem; for instance, the FEVER \cite{thorne-etal-2018-fever} dataset for fact-checking can be regarded as an NLI problem in terms of identifying the relationship between a statement and supporting evidence.

However, driven by the widespread observations that previous popular NLI datasets contain shortcuts \cite{tsuchiya-2018-performance,gururangan-etal-2018-annotation}, multiple works \cite{nie2019adversarial,glockner-etal-2018-breaking,naik-etal-2018-stress,geiger2018stress,yanaka2019help,saha2020conjnli} developed "stress tests" to benchmark specific linguistic phenomena. 

For instance, \citet{glockner-etal-2018-breaking} proposed a simple test set based on SNLI \cite{snli:emnlp2015} that involves changing a single word in the premise sentences. In this setting, performance is substantially worse than the original SNLI test set, indicating the presence of spurious correlations in the training dataset construction. \citet{naik-etal-2018-stress} proposed an NLI Stress Test by quantifying the lexical phenomena (e.g., presence of antonyms, numerical reasoning) behind common model errors in MultiNLI \cite{multi-nli}. Their proposed stress test involved constructing artificial examples that exacerbate common sources of model error, showcasing that some models have markedly reduced performance. \citet{nie2019adversarial} proposed a new benchmark called Adversarial NLI (ANLI), which leverages an interactive human-and-model-in-the-loop procedure to collect hard examples for natural language inference, obtaining a challenging stress test for current models. 

\begin{figure*}[hbt!]
    \centering
    \includegraphics[width=1.0\textwidth]{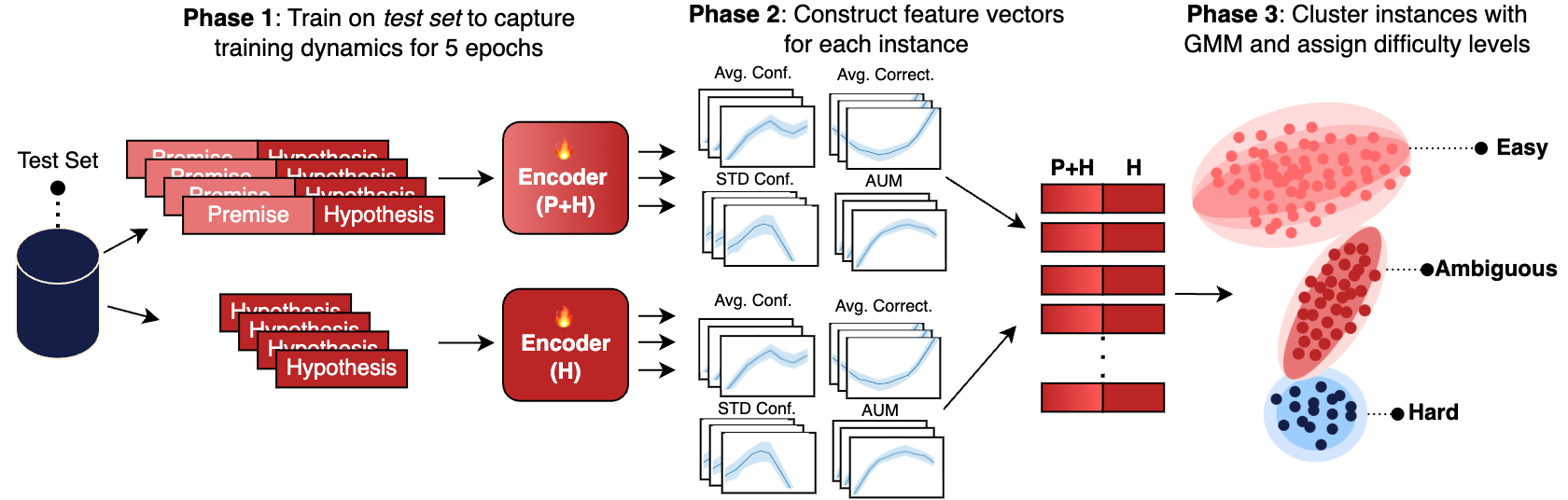}
    \caption{Overall diagram of our method to automatically construct a challenging test set for NLI. }
    \label{fig:diagram}
\end{figure*}

\citet{geiger2018stress} constructed a dataset of artificially built sentences based on first-order logic, fixing the sentence structure and only varying words corresponding to parts of speech at predefined positions. Such a dataset comprises unrealistic sentences but provides insight into the (lack-of) expressive power of certain neural architectures. Likewise, \citet{yanaka2019help} proposed the evaluation of monotonicity reasoning for NLI by constructing a dataset through curating and manipulating sentence pairs from the Parallel Meaning Bank \cite{abzianidze-etal-2017-parallel}. \citet{saha2020conjnli} identified the lack of conjunctive reasoning examples in current NLI test sets and estimated that around 72\% of sentence pairs in SNLI have conjunctions unchanged between premise and hypothesis. The authors proposed CONJNLI, a stress test composed of conjunctive sentence pairs collected automatically from Wikipedia and manually verified. 

In contrast to previous work, we propose a method to characterize the test set into multiple difficulty levels by using training dynamics of neural networks \cite{swayamdipta2020dataset,pleiss2020identifying}, thereby isolating easy and spurious examples and keeping only challenging pairs. Our method is general, model-agnostic, utilizes existing dataset samples (avoiding unrealistic artificial sentence pairs), is easily extensible to other datasets, and does not require manual verification of human annotators. Previous approaches \cite{naik-etal-2018-stress,saha2020conjnli} aim to develop a stress test for NLI by amplifying spurious correlations and evaluating model performance under various extreme conditions. Our goal is to minimize spurious correlations in existing benchmarks to gain a more realistic sense of performance under challenging real-world examples.

\section{Test Set Characterization}
\label{sec:method}

Our goal is to generalize the Data Maps proposed by \citet{swayamdipta2020dataset} to characterize the test set. \citet{swayamdipta2020dataset} proposed an approach to gauge the contribution of each training sample in a dataset by analyzing training dynamics (variability, average confidence of the gold label, and average correctness) across training for a fixed amount of epochs. After training, each example is split into one of three categories (i.e., \textit{easy-to-learn}, \textit{ambiguous} or \textit{hard-to-learn}) using a fixed percentile threshold on one of the features. For example, instances regarded as \textit{ambiguous} are examples for which the variability across 5 epochs is in the top 33\% percentiles, disregarding other measures. 

We aim to extend and generalize Data Maps by employing a Gaussian Mixture Model (GMM) \cite{Reynolds2009GaussianMM} to learn the best fitting distribution of data difficulty levels, thus avoiding fixed thresholds. Unlike other clustering techniques, such as KMeans, which outputs spherical clusters and disregards cluster variance, we chose a GMM as a more flexible clustering method. Figure \ref{fig:diagram} showcases the general methodology used in this work. We first characterize the test set by training two separate models with both premise and hypothesis (P+H), and hypothesis only (H); second, we gather 8 measures of training dynamics for each instance and cluster them to obtain three difficulty levels (4 for P+H and 4 for H). We found that difficulty levels simultaneously align with measures of spurious correlations and model performance.

In contrast to the initial approach of \citet{swayamdipta2020dataset}, we include 6 additional features for a more informative characterization across training. In our scenario focused on NLI in particular, we gather statistics for training dynamics across two types of settings: normal training (i.e., training with P + H) and hypothesis-only (H). Models trained only with the hypothesis have been shown to produce unreasonably high results \cite{poliak-etal-2018-hypothesis,liu2020hyponli}, mostly due to artifacts in dataset construction. These insights enabled us to gather statistics about such examples and improve data characterization through more diverse features for each instance. Different from \citet{swayamdipta2020dataset} who focused on characterizing the \textit{training} set for increasing data quality, we aim to construct a more challenging \textit{test set} automatically. 

As such, in order to construct a data map of the test set, we trained a model for $E$ epochs on the \textit{test set} using premise and hypothesis and, separately, using only the hypothesis to gather training dynamics for each example in the test set. Let $D_{test} = \{(x, y^*)_i\}^N_{i = 1}$ be a test dataset containing $N$ instances, where, in our case, $x_i$ is comprised of a premise + hypothesis pair or only a hypothesis. We compute the following measures across training an Encoder model in both scenarios (P + H and H) for each example $x_i$: confidence ($\hat{\mu}_i$), variability ($\hat{\sigma}_i$), correctness ($\hat{c}_i$) and Area Under Margin (${\textnormal{\textit{AUM}}}_i$). 

\begin{equation}
    \hat{\mu}_i = \frac{1}{E} \sum^{E}_{e = 1}p_{\boldsymbol{\theta}^{(e)}}(y^*_i | \boldsymbol{x_{i}})
    \label{eq:confidence}
\end{equation}

\begin{equation}
    \hat{\sigma}_i = \sqrt{\frac{\sum^{E}_{e = 1}(p_{\boldsymbol{\theta}^{(e)}}(y^*_i | \boldsymbol{x_{i}}) - \hat{\mu}_i)}{E}}
    \label{eq:variability}
\end{equation}

\begin{equation}
    \hat{c}_i = \frac{1}{E} \sum^{E}_{e = 1} [\operatorname*{argmax}(p_{\boldsymbol{\theta}^{(e)}}(x_i)) = y^*_i]
    \label{eq:corectness}
\end{equation}

where $p_{\boldsymbol{\theta}^{(e)}}$ corresponds to the model's probability during training at epoch $e$. Following \citet{swayamdipta2020dataset}, we compute confidence, variability, and correctness concerning the correct label $y^*_i$. Furthermore, we compute AUM \cite{pleiss2020identifying}, which was initially proposed to identify mislabeled examples but yielded a similar type of characterization as Data Maps. We include AUM as an additional measure of instance correctness/learnability. Let $z^{(e)}_y (x_i)$ be the logit (pre-softmax) for class $y$ of the model at epoch $e$, given an instance $x_i$. The area under margin (AUM or average margin) of $x_i$ is computed as:

\begin{equation}
    {\textnormal{\textit{AUM}}}_i = \frac{1}{E} \sum^{E}_{e = 1} (z^{(e)}_{y^*_i}(x_i) - \operatorname*{max}_{(y \neq y^*_i)} z^{(e)}_y (x_i))
    \label{eq:AUM}
\end{equation}

In all our experiments, we first fine-tune pretrained RoBERTa models \cite{liu2019roberta}, followed by DeBERTa \cite{he2022debertav3} models due to their established high performance on a wide set of tasks. In our formulation, any other encoder would yield similar results, as this method is based only on the final classification output and not model internals. However, characterization based on final logits and class confidences is affected by how calibrated the models' predictions are \cite{guo2017calibration}, as poorly calibrated models have lower logit variance across classes. For our scope, we are interested in identifying and separating spurious correlations in NLI benchmarks and not in benchmarking different classifiers for this task. We explore the impact of the underlying encoder in Section \ref{sec:classifier}.

In Table \ref{tab:roberta-results}, we show the results of our RoBERTa models trained on SNLI, MultiNLI, and FEVER on different configurations of training/testing splits and using both the premise and the hypothesis or only the hypothesis. Our reproduction of results is on par with other works. For completeness, we also show results where the model is trained on the test set, but note that the purpose is only to gather training dynamics and not directly use it as a classifier. The model trained on only the hypothesis obtains 71\% accuracy on SNLI and 61\% accuracy on MultiNLI, while random chance performance is 33\%. These results strongly point toward spurious correlations and annotation artifacts on both datasets \cite{tsuchiya-2018-performance,gururangan-etal-2018-annotation,poliak-etal-2018-hypothesis,liu2020hyponli}.  In the case of FEVER, the hypothesis-only model achieved close to random-chance, indicating less spurious correlations found in the hypothesis.

\begin{table}[hbt!]
    \centering
    \resizebox{1.0\linewidth}{!}{
    \begin{tabular}{lll|cc}
        \textbf{Dataset} & \textbf{Train Split} & \textbf{Test Split} & \textbf{Accuracy (P + H)} & \textbf{Accuracy (H)} \\
        \toprule
        \multirow{2}{*}{\textbf{SNLI}} &  Train & Test & 0.9178 & 0.7170 \\
        & Test & Test & 0.9799 & 0.8764 \\
        \midrule
        \multirow{2}{*}{\textbf{MultiNLI}} & Train & Val & 0.8773 & 0.6142 \\
        & Val & Val & 0.9841 & 0.8612 \\
        \midrule
        \multirow{2}{*}{\textbf{FEVER}} & Train & Dev & 0.7702  & 0.3822 \\
        & Dev & Dev & 0.9459 & 0.7137 \\
    \end{tabular}
    }
    \caption{Results for RoBERTa on SNLI, MultiNLI and FEVER under various train/test splits and input types.}
    \label{tab:roberta-results}
\end{table}

After training two classifiers (P + H and H) on each dataset's test set, we construct a feature vector $f_i$ describing the training dynamics of an instance $i$ by concatenating the training dynamics of each sample in both settings:

\begin{equation}
    \begin{split}
    f^{(P+H)}_i &= [\hat{\mu}^{(P + H)}_i, \hat{\sigma}^{(P + H)}_i, \hat{c}^{(P + H)}_i, {\textnormal{\textit{AUM}}}^{(P + H)}_i] \\
    f^{(H)}_i &= [\hat{\mu}^{(H)}_i, \hat{\sigma}^{(H)}_i, \hat{c}^{(H)}_i, {\textnormal{\textit{AUM}}}^{(H)}_i] \\
    f_i &= f^{(P+H)}_i \mathbin\Vert f^{(H)}_i
    \end{split}
    \label{eq:features}
\end{equation}

\begin{figure*}[hbt!]
    \centering
    \includegraphics[width=0.8\textwidth]{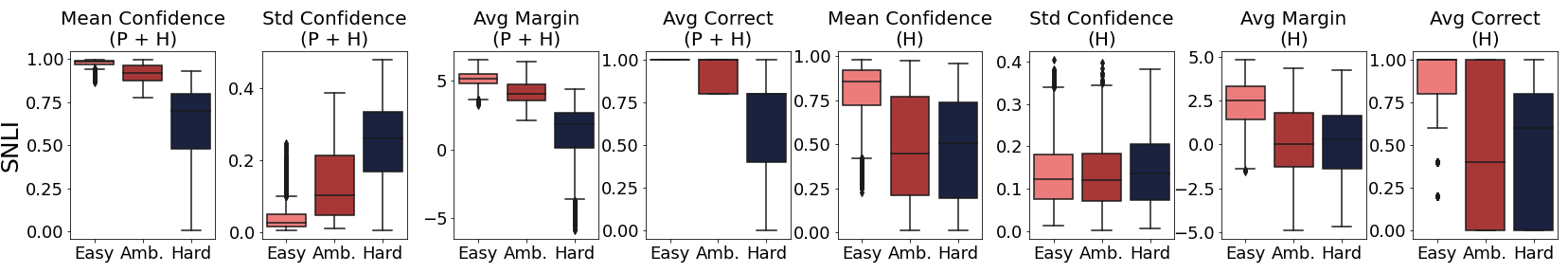}
    \includegraphics[width=0.8\textwidth]{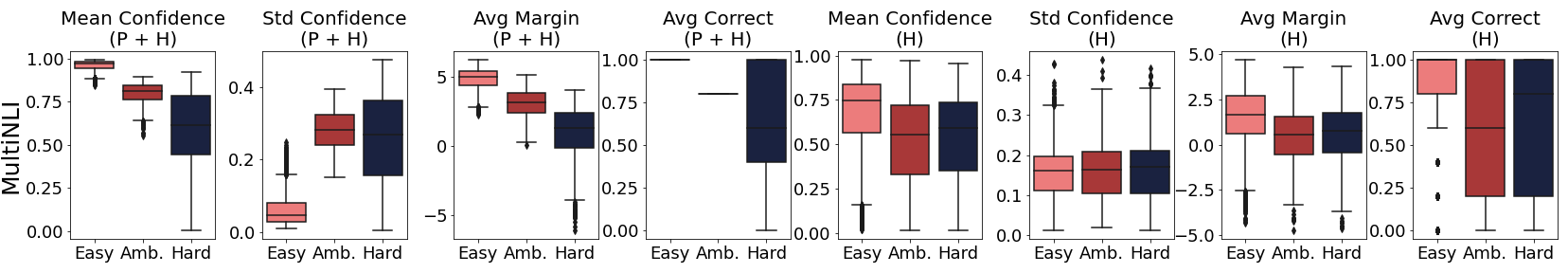}
    \includegraphics[width=0.8\textwidth]{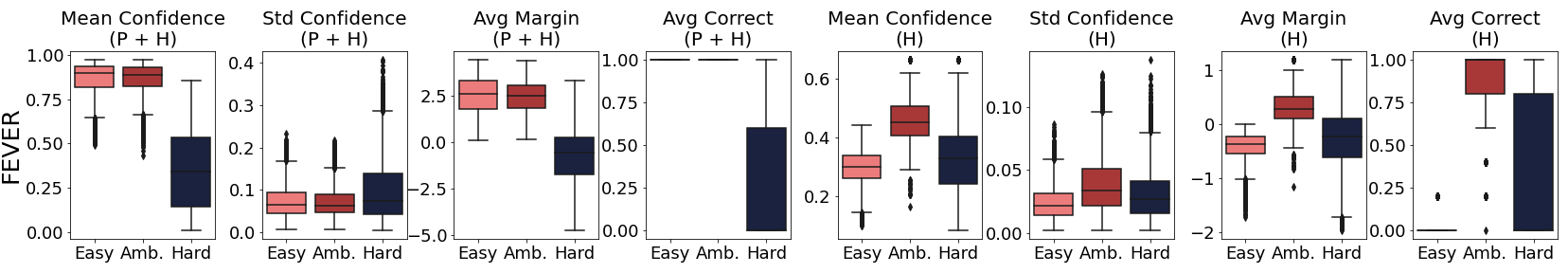}
    \caption{Distributions of feature values across difficulty levels for the test set for SNLI (top), MultiNLI (middle), and FEVER (bottom). In addition to features explored in Data Maps \cite{swayamdipta2020dataset}, we also incorporated the Average Margin \cite{pleiss2020identifying} and included training dynamics across a model trained only on the hypothesis.}
    \label{fig:snli-mnli-features}
\end{figure*}

Using the feature vectors $\{f_i\}^{N}_{i=1}$, we cluster the test set using a Gaussian Mixture Model into three clusters. Feature vectors are normalized with standard scaling by subtracting the mean and dividing by the standard deviation of each feature. The clusters are ranked according to the intra-cluster average confidence $\hat{\mu}^{(P + H)}$, and we interpret them as belonging to three difficulty levels, following the terminology introduced by \citet{swayamdipta2020dataset}: \textit{easy}, \textit{ambiguous} and \textit{hard}, in decreasing order of the average intra-cluster confidence $\overline{\hat{\mu}}^{(P + H)}$. See Appendix \ref{sec:appendix} for a high-level overview of our algorithm.

Figure \ref{fig:snli-mnli-features} depicts the distribution of features across difficulty levels for both datasets. While each type of feature captures different aspects of the learnability of an instance, their combination offers a more diverse view of the learning dynamics during training. In the case of both SNLI and MultiNLI, harder examples have consistently lower average margin and more variability of the correct class. The effect is not as pronounced in a hypothesis-only setting; however, there is a clear delimitation of easy examples for average margin and confidence, indicating potential annotation artifacts.  For FEVER, since the dataset has reduced spurious correlations, the identified splits correspond to difficult-to-learn examples, not necessarily examples with annotation artifacts.

In the interest of quantifying the number of spurious correlations found in the test set, we follow \citet{naik-etal-2018-stress} and track several measures that correspond to either shallow statistics between the premise and the hypothesis, or the presence of negations or misspelled words. Table \ref{tab:heuristics} presents the heuristics implemented in our work. We automatically compute each measure, avoiding time-consuming manual annotations.

\begin{table}[hbt!]
    \centering
    \resizebox{1.0\linewidth}{!}{
    \begin{tabular}{l|p{6cm}}
        \textbf{Name} & \textbf{Explanation} \\
        \toprule
        Word Overlap & Number of common words between the premise and hypothesis, normalized by sentence length \\
        
        Number of Antonyms & Number of antonyms of each of the words in the premise contained in hypothesis, based on WordNet \cite{Fellbaum1998}, normalized by sentence length. \\
        
        Length Mismatch & Difference in length between premise and hypothesis, normalized by sentence length\\
        
        Misspelled Words & Total number of misspelled words using a spellchecker in the premise and hypothesis, normalized by sentence length. \\
        
        Contains Negation & Boolean flag if either the premise of hypothesis contains a negation word (e.g., \textit{no, not, never, none}) \\
    \end{tabular}
    }
    \caption{Heuristic measures of spurious correlations used, similar to the categories by \citet{naik-etal-2018-stress}.}
    \label{tab:heuristics}
\end{table}
\begin{table}[hbt!]
    \centering
    \resizebox{1.0\linewidth}{!}{
    \begin{tabular}{cc|cp{1.5cm}p{1.5cm}}
       \textbf{Dataset} & \textbf{Split} & \textbf{Fraction of total} & \textbf{Accuracy (P + H)} &  \textbf{Accuracy (H)} \\
       \toprule
        \multirow{3}{*}{\textbf{SNLI}} & Easy & 0.70 \textit{(6889 / 9824)} & 0.97 & 0.82 \\
         & Amb. & 0.17 \textit{(1725 / 9824)} & 0.89 & 0.46 \\
         & Hard & 0.12 \textit{(1210 / 9824)} & 0.56 & 0.38 \\
        \midrule
        \multirow{3}{*}{\textbf{MultiNLI}} & Easy & 0.75 \textit{(7381 / 9815)} & 0.94 & 0.67 \\
         & Amb. & 0.14 \textit{(1420 / 9815) }& 0.75 & 0.43 \\
         & Hard & 0.10 \textit{(1014 / 9815) }& 0.53 & 0.39 \\
         \midrule
        \multirow{3}{*}{\textbf{FEVER}} & Easy & 0.50 \textit{(10083 / 19998)} & 0.95 & 0.43 \\
         & Amb. & 0.24 \textit{(4903 / 19998) }& 0.88 & 0.49 \\
         & Hard & 0.25 \textit{(5012 / 19998) }& 0.29 & 0.16 \\
    \end{tabular}
    }
    \caption{Performance of RoBERTa models trained on the training set and evaluated on different splits of our stress test.}
    \label{tab:accuracy-per-split}
\end{table}

Training the pretrained models for dataset characterization was performed for 5 epochs each, with a batch size of 32, using the Adam optimizer with a learning rate of $10^{-5}$ following a linear decay schedule with warm-up. We used mixed precision for all training runs.

\section{Results \& Discussion}

\begin{figure*}[hbt!]
    \centering
    \includegraphics[width=0.8\linewidth]{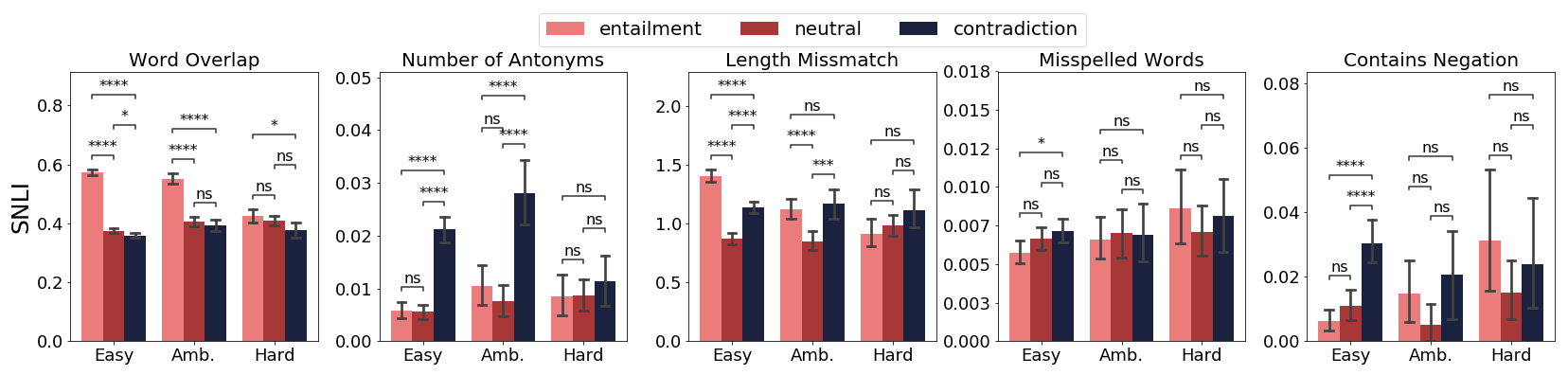}
    \includegraphics[width=0.8\linewidth]{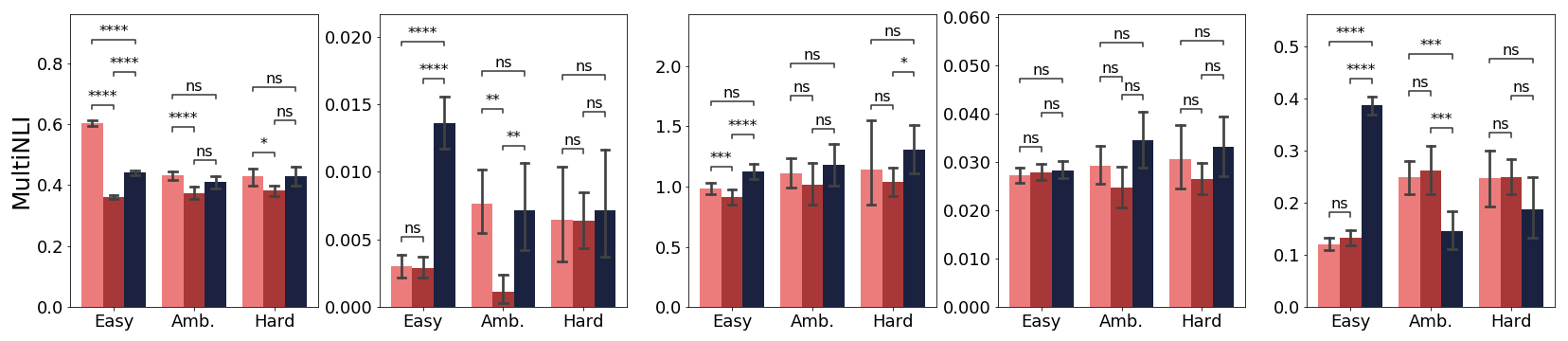}
    \includegraphics[width=0.8\linewidth]{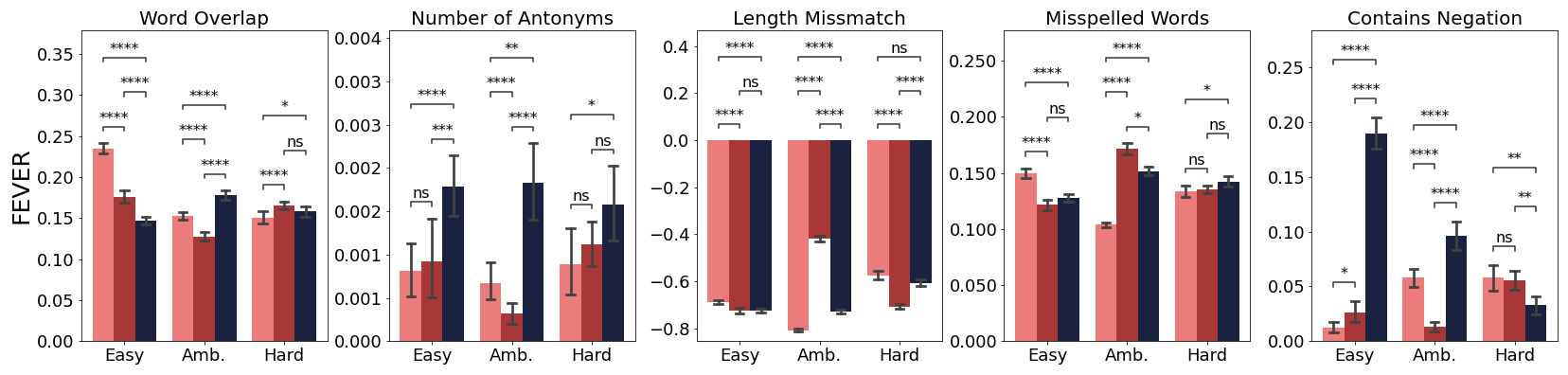}
    \caption{Distributions of the measures of spurious correlations for each level (\textit{easy}, \textit{ambiguous}, \textit{hard}) across the three labels (\textit{entailment}, \textit{neutral}, \textit{contradiction}) for SNLI (top), MultiNLI (middle) and FEVER (bottom).}
    \label{fig:stats-heuristics}
\end{figure*}

Table \ref{tab:accuracy-per-split} shows the performance of a RoBERTa model trained on each dataset's training set and evaluated on our stress test after characterization using training dynamics. Easier instances have more examples annotated with "contradiction" and "entailment", while harder instances have more examples annotated with "neutral". Performance monotonously degrades upon increasing difficulty levels, reaching 56\% accuracy on SNLI-hard and 53\% accuracy on MultiNLI-hard. Performance on the \textit{easy} split for both datasets is considerably higher compared to the global accuracy with all splits combined. Furthermore, the accuracy of a model trained using only the hypothesis degrades to almost random chance on harder splits, indicating that the \textit{hard} split has fewer annotation artifacts. Compared to \citet{swayamdipta2020dataset}, the difficulty levels are not equal in size; the majority ($\sim$70\%) of samples belongs to the \textit{easy} category, while only around 10\% are characterized as being \textit{hard}. For FEVER, the performance degradation in the hard split is more dramatic: 29\% accuracy for \textit{hard} compared to 88\% for ambiguous, indicating truly difficult examples for the model.

\begin{figure}[hbt!]
    \centering
    \includegraphics[width=1.0\linewidth]{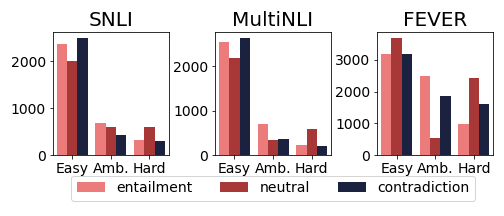}
    \caption{Counts for each class in SNLI, MultiNLI, and FEVER, according to each difficulty level.}
    \label{fig:hardness-counts}
\end{figure}

Even though the \textit{hard} split is relatively small, the subsample is challenging for current models, as it comprises instances with fewer spurious correlations between premise and hypothesis. Fewer annotation artifacts enable fewer "correct" predictions from linguistic patterns present only in the hypothesis. In Figure \ref{fig:hardness-counts}, we show per-class counts relative to the difficulty levels for both datasets. It is the case that easier samples contain more contradictions and entailments, prone to linguistic commonalities (word overlap, presence of antonyms). Thus, the \textit{hard} split has more neutral instances. 

Through manual inspection, we found that for SNLI, the \textit{easy} splits contain unrelated sentences which are sometimes annotated incorrectly as Contradiction (e.g., "a woman running in the park" versus "a man cooking at home" - two unrelated sentences annotated as contradicting). The model learns this pattern and incorrectly predicts Contradiction on some Neutral pairs (e.g., "... girls chatting on the stairwell" versus "girls are at school"). For MultiNLI, we found that the easy split usually aligns with simple sentence negations (e.g., "it gets it" versus "it doesn't get it") or paraphrasing ("I guess history repeats itself" versus "history certainly doesn't repeat"). These observations strongly point towards spurious correlations between premise and hypothesis, making the sentences easier to classify correctly. We provide selected examples in the Appendix \ref{sec:appendix}. The \textit{ambiguous} and \textit{hard} splits in both datasets contain increasingly more subtle cues, with little overlap in words between premise and hypothesis (e.g., "standing on a tree log" versus "crossing the stream" / "wouldn't have mattered" versus "would have gotten worse"), having more natural and challenging sentence pairs.

Across SNLI, MultiNLI, and FEVER, we tracked the average amount of each measure between difficulty levels and classes (see Figure \ref{fig:stats-heuristics}). To rigorously test the difference between the classes at various difficulty levels, we perform a non-parametric two-sided Mann-Whitney-U test \cite{mann1947test} with Bonferroni correction to test for statistical significant differences\footnote{Significance thresholds: Not Significant (ns): $.05 < p$, *:~$.01 \le p \leq .05$, **: $.001 \le p \leq .01$, ***: $p \leq .001$,}. 

We found no evidence for the presence of spurious correlations (Mann-Whitney's U test $p > .05$)
in the \textit{hard} split between the three classes. Some measures are more associated with certain classes. For example, instances annotated with Contradiction have a disproportionate amount of antonyms between premise and hypothesis in the \textit{easy} and \textit{ambiguous} splits. Similarly, negation is more present in the Contradiction class for easy splits. For instances annotated with Entailment, word overlap is present significantly in \textit{easy} splits. Between SNLI and MultiNLI, MultiNLI has a disproportionately large amount of negations compared to SNLI. For both SNLI and MultiNLI, our method yields little to no significant differences between classes in the \textit{hard} split across the spurious correlation measures. For FEVER, measures such as the presence of negations, number of antonyms, and word overlap are reduced across difficulty levels. Note that FEVER includes a small statement as the premise and a long text extract containing evidence as the hypothesis, which makes the length mismatch negative.

\subsection{Impact of the Underlying Encoder}
\label{sec:classifier}

To show that our method is model-agnostic, we further provide a comparison between the dataset characterizations obtained by RoBERTa and DeBERTa. Table \ref{tab:classifier-comparison} showcases the accuracies of the two models on each others' data characterizations. The difficulty splits are maintained cross-model.

\begin{table}[hbt!]
    \centering
    \resizebox{1.0\linewidth}{!}{
    \begin{tabular}{cc|ccc}
        &  & & \textbf{Source: RoBERTa} & \textbf{Source: DeBERTa}\\
       \textbf{Split} & \textbf{Dataset} & \textbf{Target Model} & \textbf{Accuracy} & \textbf{Accuracy}\\
       \toprule
      \multirow{6}{*}{\textbf{Easy}} & \multirow{2}{*}{\textbf{SNLI}} & RoBERTa & 0.9779 & 0.9462 \\
       & & DeBERTa & 0.9792 & 0.9624 \\
       \cmidrule{2-5}
       & \multirow{2}{*}{\textbf{MultiNLI}} & RoBERTa & 0.9470 & 0.9502 \\
       &  & DeBERTa & 0.9545 & 0.9567 \\
       \cmidrule{2-5}
      & \multirow{2}{*}{\textbf{FEVER}} & RoBERTa &  0.9101  & 0.9346  \\
       &  & DeBERTa & 0.8967 & 0.9501  \\
       \midrule
       \multirow{6}{*}{\textbf{Ambiguous}} & \multirow{2}{*}{\textbf{SNLI}} & RoBERTa & 0.8916 & 0.9802 \\
       & & DeBERTa & 0.9003 & 0.9881 \\
        \cline{2-5}
       & \multirow{2}{*}{\textbf{MultiNLI}} & RoBERTa & 0.7577 & 0.9086 \\
       &  & DeBERTa & 0.7746 & 0.9543 \\
       \cmidrule{2-5}
       & \multirow{2}{*}{\textbf{FEVER}} & RoBERTa & 0.9532  & 0.8697 \\
       &  & DeBERTa & 0.9470 & 0.8697 \\
       \midrule
      \multirow{6}{*}{\textbf{Hard}} & \multirow{2}{*}{\textbf{SNLI}} & RoBERTa & 0.5678 & 0.6337\\
       &  & DeBERTa & 0.6446 & 0.6437 \\
      \cmidrule{2-5}
       & \multirow{2}{*}{\textbf{MultiNLI}} & RoBERTa & 0.5375 & 0.7497 \\
       &  & DeBERTa & 0.6460 & 0.7585 \\
       \cmidrule{2-5}
       & \multirow{2}{*}{\textbf{FEVER}} & RoBERTa & 0.3444 & 0.2913 \\
       &  & DeBERTa & 0.4089 & 0.2977 \\
    \end{tabular}
    }
    \caption{Comparison between RoBERTa and DeBERTa accuracy on each difficulty level, across models.}
    \label{tab:classifier-comparison}
\end{table}

\begin{figure}[hbt!]
    \centering
    \includegraphics[width=1.0\linewidth]{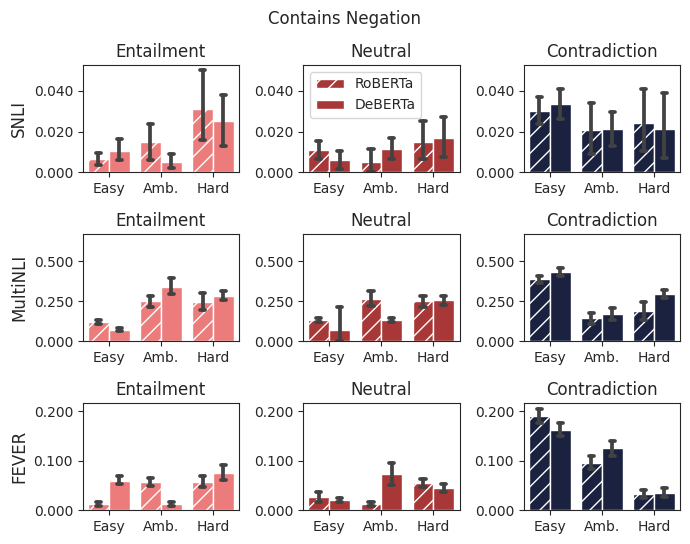}
    \caption{Comparison between the characterizations obtained by RoBERTa and DeBERTa on the "Contains Negation" heuristic measure.}
    \label{fig:comparison-contains-negation}
\end{figure}

Across datasets and difficulty levels, the performance sharply drops for the \textit{"hard"} split for both models. DeBERTa achieved higher accuracy for \textit{"hard"} set, most likely due to better overall performance compared to RoBERTa. In Figure \ref{fig:comparison-contains-negation}, we show that overall heuristic values for "Contains Negation" are maintained across both models. Extended results for are presented in Appendix \ref{sec:appendix}.

Our proposed methodology is general and independent of the underlying encoder model since we process training dynamics computed from raw logit scores. This characterization procedure may be adapted to using Large Language Models (LLMs) \cite{lee2023can} in a zero-shot classification setting by manipulating the log-likelihood for the tokens of the correct classes. However, using LLMs requires a different approach than the one presented here since the networks are usually used without further training, in an in-context-learning manner \cite{dong2022survey}. Furthermore, even if the LLMs are fine-tuned \cite{hu2021lora}, it is not straightforward how the logits of each of the three classes are tracked across training. We leave this approach for future work.

\subsection{Training Set Characterization}
\label{sec:exp}

\begin{table}[hbt!]
    \centering
    \resizebox{1.0\linewidth}{!}{
    \begin{tabular}{p{1.2cm}ll|p{1.5cm}p{1.5cm}p{1.5cm}p{1.5cm}p{1.5cm}cc}
        & & \multicolumn{7}{c}{\textbf{Stress Test by \citet{naik-etal-2018-stress}}}\\
        \textbf{Method} & \textbf{Splits} & \textbf{\% SNLI} &  \textbf{Word} \newline \textbf{Overlap} & \textbf{Spelling} \newline \textbf{Error} & \textbf{Numerical} \newline \textbf{Reasoning} &  \textbf{Negation} & \textbf{Length} \newline \textbf{Mismatch} & \textbf{Antonym} & \textbf{Overall} \\
        \toprule
        DataMaps  & Amb.  & 33\% & 0.6792 & 0.7202 & 0.3745 & 0.5388 & 0.7176 & 0.6237 & 0.6662\\
        AUM & Amb.  & 33\% & 0.6818 & 0.6904 & 0.2896 & 0.4837 & 0.7134 & \textbf{0.7417} & 0.6416\\
        \midrule
        \multirow{6}{*}{\textbf{Ours}} & Easy                & 53\%  & 0.6720 &       0.6713 &            \underline{0.4287} &  0.4622 &        0.6870 &  0.4446 & 0.6246 \\
        & Easy+Amb    & 86\%  & 0.7231 &        \textbf{0.7500} &            \textbf{0.4452} &  \underline{0.6124} &        \textbf{0.7578} &  0.6582 & \underline{0.7083} \\
        & Amb           & 33\%  & \underline{0.7264} &       \underline{0.7414} &            0.4158 &  \textbf{0.6666} &        \underline{0.7491} &  0.6515 & \textbf{0.7093} \\
        & AmbHard    & 46\%  & \textbf{0.7302} &        0.7305 &            0.3761 &  0.5533 &        0.7471 &  \underline{0.6619} & 0.6860\\
        & Easy+Hard         & 66\%  & 0.6714 &       0.6815 &            0.3803 &  0.5617 &        0.6993 &  0.4971 & 0.6442\\
        & Hard                & 13\%  & 0.3191 &       0.3310 &            0.3304 &  0.3184 &        0.3189 &     0.0 & 0.3177\\
        \midrule
        & All        & 100\% & 0.7252 & 0.7559 & 0.2794 & 0.5923 & 0.7771 & 0.7268 & 0.7037\\
    \end{tabular}
    }
    \caption{Results for a RoBERTa model trained on SNLI in various configurations and evaluated on the stress test by \citet{naik-etal-2018-stress} based on MultiNLI. The best results are \textbf{bold}, while the second best are \underline{underlined}.}
    \label{tab:snli-their-stress-test-results}
\end{table}
\begin{table}[hbt!]
    \centering
    \resizebox{1.0\linewidth}{!}{
    \begin{tabular}{p{1.2cm}ll|p{1.5cm}p{1.5cm}p{1.5cm}p{1.5cm}p{1.5cm}cc}
        & & \multicolumn{6}{c}{\textbf{Stress Test by \citet{naik-etal-2018-stress}}}\\
        \textbf{Method} & \textbf{Splits} & \textbf{\% MultiNLI} & \textbf{Word} \newline \textbf{Overlap} & \textbf{Spelling} \newline \textbf{Error} & \textbf{Numerical} \newline \textbf{Reasoning} &  \textbf{Negation} & \textbf{Length} \newline \textbf{Mismatch} & \textbf{Antonym} & \textbf{Overall} \\
        \toprule
        DataMaps & Amb.    & 33\% & \underline{0.7046} & 0.7938 & 0.4065 & 0.5469 & 0.8197 & 0.5712 & 0.7222\\
        AUM & Amb.          & 33\% & 0.6428 & 0.7998 & 0.2888 & \underline{0.5638} & 0.8254 & 0.5129 & 0.7115\\
        \midrule
        \multirow{6}{*}{\textbf{Ours}} & Easy & 41\% & 0.6885 &       0.7978 &  0.3391 &  0.5514 &        0.8210 &  0.5274 & 0.7179\\
        & Easy+Amb.   & 84\% & 0.6601 &       \underline{0.8274} &            \textbf{0.4777} &  \textbf{0.5764} &        \underline{0.8467} &  \textbf{0.6330} & \underline{0.7459} \\
        & Amb.          & 43\% &  0.6283 &       0.8235 &            0.4602 &  0.5578 &        0.8421 &  0.5742 & 0.7337 \\
        & Amb+Hard   & 59\% & 0.6961 &       \textbf{0.8255} &            \underline{0.4651} &  0.5628 &         \textbf{0.8476} &  \underline{0.6288} & \textbf{0.7474} \\
        & Easy+Hard   & 56\% & \textbf{0.7170} &       0.8032 &            0.2948 &  0.5607 &        0.8258 &  0.5062 & 0.7237\\
        & Hard & 15\%   & 0.4525 &       0.4966 &            0.2705 &  0.4589 &        0.4914 &  0.2506 &   0.4656 \\
        \midrule
        & All & 100\% & 0.6701 & 0.8308 & 0.5380 & 0.5654 & 0.8500 & 0.6312 & 0.7511\\
    \end{tabular}
    }
    \caption{Results for a RoBERTa model trained on MultiNLI in various configurations and evaluated on the stress test proposed by \citet{naik-etal-2018-stress} based on MultiNLI. The best results are \textbf{bold}, while the second best are \underline{underlined}.}
    \label{tab:multinli-their-stress-test-results}
\end{table}

Our method provides a more challenging test set devoid of shortcuts and spurious correlations. Further, we explore the possibility of using this approach to improve data quality for training NLI models. We employ the same algorithm to characterize the \textit{training sets} for SNLI and MultiNLI and train a RoBERTa model on the different resulting combinations of difficulty levels. Under each configuration, the model is trained for 10 epochs with early stopping on the validation set loss, with a learning rate of $10^{-5}$ following a linear decay schedule with a warm-up. 

We evaluate each model on the stress test proposed by \citet{naik-etal-2018-stress} that is based on MultiNLI. However, we emphasize that the stress test of \citet{naik-etal-2018-stress} is designed to unrealistically amplify spurious correlations to gauge the model performance under various extreme conditions, in contrast to our method, which eliminates linguistic shortcuts while mimicking real-world examples. In Tables \ref{tab:snli-their-stress-test-results} and \ref{tab:multinli-their-stress-test-results}, we show the performance on the dataset proposed by \citet{naik-etal-2018-stress} for RoBERTa models trained on SNLI and MultiNLI. The authors provided metadata for each instance that allows fine-grained evaluation under different linguistic reasoning phenomena.

We compared our approach with Data Maps \cite{swayamdipta2020dataset} and Area Under Margin \cite{pleiss2020identifying}, two popular methods for training set characterizations using training dynamics. For Data Maps, we select \textit{ambiguous} examples by keeping the instances where average variability is in the top 66\% percentile. For AUM, while the authors did not explicitly propose a threshold for characterizing each instance, we follow a similar approach to Data Maps by considering \textit{ambiguous} examples to have an average margin between the 33\% and 66\% percentiles. Our method outperforms Data Maps and AUM across the majority of settings and, in some cases, outperforms a model trained on the full dataset while using a smaller amount of data but of higher quality. This indicates that our method is a viable alternative to AUM or Data Maps for increasing dataset quality.


\section{Conclusions}
\label{sec:conc}

Our method highlights significant shortcomings in widely used NLI evaluation datasets (SNLI and MultiNLI) due to spurious correlations in the annotation process. To address these issues, we proposed an automatic method for constructing more challenging test sets, effectively filtering out problematic instances and providing a more realistic measure of model performance. Our approach, which categorizes examples in increasing difficulty levels using a wide range of training dynamics features, enhances evaluation reliability and offers insights into underlying challenges in NLI. Importantly, our methodology is general and model-agnostic, and can be applied across different datasets and models, promising improved evaluation practices in NLP. 

Furthermore, we provided evidence that our method can obtain a challenging test set even if the dataset has fewer annotation artifacts; we characterized FEVER, a fact-checking dataset repurposed for NLI, and showed that the identified hard split is a highly challenging subset of the dataset. By aggressively filtering uninformative examples, we show that comparable model performance can be achieved with significantly reduced data requirements. Our work contributes to advancing NLI evaluation standards, fostering the development of more robust NLU models.

\section*{Limitations}
\label{sec:lim}

Our method is unsuitable for automatically identifying mislabeled examples in a dataset. While it does incorporate measures such as Area Under Margin \cite{pleiss2020identifying}, designed with this purpose, proper manual verification is needed to increase annotation quality. 

\section*{Acknowledgements}
\label{sec:ack}

The work of Adrian Cosma was supported by a mobility project of the Romanian Ministery of Research, Innovation and Digitization, CNCS - UEFISCDI, project number PN-IV-P2-2.2-MC-2024-0641, within PNCDI IV. The work of Stefan Ruseti was supported by a mobility project of the Romanian Ministery of Research, Innovation and Digitization, CNCS - UEFISCDI, project number PN-IV-P2-2.2-MC-2024-0585, within PNCDI IV.
The work was also supported by a grant from the National Science Foundation NSF/IIS \#2107518. 
\bibliography{refs}
\bibliographystyle{acl_natbib}

\appendix

\section{Appendix}
\label{sec:appendix}

\subsection{Algorithm}

We present the high-level overview of our methodology in Algorithm \ref{alg:test-set-construction}.

\begin{algorithm}[!hbt]
    \caption{Pseudo-code for the construction of a stress test based on training dynamics.} 
    \label{alg:test-set-construction}
    \begin{algorithmic}
     \scriptsize
        \Require $D_{test}$ - the target test set \\

        \State Train encoder model $M_{\theta}^{(P+H)}$ on  $D_{test}$ using the premise and hypothesis for 5 epochs, tracking training dynamics
        \State Compute $\hat{\mu}^{(P+H)}, \hat{\sigma}^{(P+H)}, \hat{c}^{(P+H)}, \text{AUM}^{(P+H)}$ (Eqs. \ref{eq:confidence},\ref{eq:variability}, \ref{eq:corectness}, \ref{eq:AUM})
        \State Construct training dynamics features $f_i^{(P+H)}$ for each instance $i$ (Eq. \ref{eq:features})
        \\
        \State Train encoder model $M_{\phi}^{(H)}$ on hypothesis for $E$ epochs, tracking training dynamics
        \State Compute $\hat{\mu}^{(H)}, \hat{\sigma}^{(H)}, \hat{c}^{(H)}, \text{AUM}^{(H)}$ (Eqs. \ref{eq:confidence},\ref{eq:variability}, \ref{eq:corectness}, \ref{eq:AUM})
        \State Construct training dynamics features $f_i^{(H)}$ for each instance $i$ (Eq. \ref{eq:features})
        \\

        \State Concatenate features vectors $f_i = f^{(P+H)}_i \mathbin\Vert f^{(H)}_i$
        
        \\
        \State Train a Gaussian Mixture Model with 3 clusters on $f_i$ 
        \State Order clusters based on average intra-cluster confidence $\overline{\hat{\mu}}$, considering "easy" (e), "ambiguous" (a) and "hard" (h) having $\overline{\hat{\mu}}^{(e)} > \overline{\hat{\mu}}^{(a)} > \overline{\hat{\mu}}^{(h)}$

        \\
        \State Split $D_{test}$ into $D_{test}^{(e)}$, $D_{test}^{(a)}$, $D_{test}^{(h)}$, based on examples corresponding to each cluster 
    
        \State \textbf{return} $D_{test}^{(e)}$, $D_{test}^{(a)}$, $D_{test}^{(h)}$
    \end{algorithmic}
\end{algorithm}
\begin{table*}[hbt!]
    \centering
    \resizebox{0.9\textwidth}{!}{
    \begin{tabular}{cc|ll|llc}
       \textbf{Dataset} & \textbf{Difficulty} & \textbf{Premise} & \textbf{Hypothesis}  & \textbf{True Label} & \textbf{Model Prediction} & \textbf{Correct?}\\
       \toprule
        \multirow{12}{*}{\textbf{SNLI}} 
        & \multirow{4}{*}{\textbf{Easy}}  & A brown dog plays in a deep pile of snow. & A brown dog plays in snow & Entailment & Entailment & \checkmark \\
        & & Woman running in a park while listening to music. & A man cooking at home. & Contradiction & Contradiction & \qmark \\
        & & Two daschunds play with a red ball & A cat in a litter box. & Contradiction & Contradiction & \qmark \\
        & & A grim looking man with sunglasses pilots a boat. & The happy pilot flies his plane. & Contradiction& Contradiction & \qmark \\
        \cmidrule{2-7}
        & \multirow{4}{*}{\textbf{Amb.}} & An older women tending to a garden. & The lady has a garden & Entailment & Entailment & \checkmark\\
        & & People are hiking up a mountain with no greenery. & The hikers have backpacks. & Neutral & Neutral & \checkmark\\
        & & A man in a suit speaking to a seated woman. & A man in a costume speaking to another man. & Contradiction & Contradiction & \checkmark \\
        & & A helmeted male airborne on a bike on a dirt road. & The man fell off his bike. & Contradiction & Contradiction & \qmark \\
        \cmidrule{2-7}
        & \multirow{4}{*}{\textbf{Hard}} & A couple is taking a break from bicycling. & a couple sit next to their bikes. & Neutral & Contradiction & \xmark \\
        & & Three kids in a forest standing on a tree log. & Children crossing stream in forest. & Neutral & Contradiction & \xmark \\
        & & A car is loaded with items on the top. & The car is a convertible.. & Contradiction & Neutral & \xmark \\
        & & 3 girls chatting and laughing on the stairwell. & Girls are at school. & Neutral & Contradiction & \xmark \\
        \midrule
        \midrule
         \multirow{12}{*}{\textbf{MultiNLI}} & \multirow{4}{*}{\textbf{Easy}} 
         & Through a friend who knows the lift boy here. & A friend knows the lift boy here. & Entailment & Entailment & \checkmark \\
         & & I guess history repeats itself, Jane. & History certainly doesn't repeat, Jane. & Contradiction & Contradiction & \checkmark \\
         & & He says men are here. & He said that the men were not here. & Contradiction & Contradiction & \checkmark \\
         &  & it gets it & It doesn't get it. & Contradiction & Contradiction & \checkmark \\
         \cmidrule{2-7}
         & \multirow{4}{*}{\textbf{Amb.}} & He slowed. & He stopped moving so quickly. & Entailment & Neutral &\xmark \\
         &  & uh high humidity & Air with increased water content. & Entailment & Entailment & \checkmark \\
         &  & I don't know all the answers, fella. & Buddy, I just can't answer all those questions. & Entailment & Neutral & \xmark\\
         & & British action wouldn't have mattered. & If Britain got involved, things would have gotten worse. &  Contradiction & Neutral & \xmark \\
         \cmidrule{2-7}
         & \multirow{4}{*}{\textbf{Hard}} & Detroit Pistons they're not as good as they were last year & Detroit Pistons played better last year & Entailment & Entailment & \checkmark \\
         & & The White House denies this. & The White House, off the record, knows it to be true. & Neutral & Contradiction & \xmark \\
         & & I'm not interested in tactics, Al. & Al is very interested in tactics. & Neutral & Contradiction & \xmark \\
         & & The four Javis children? asked Severn. & You have to ask Severn about the four Jarvis children. & Contradiction & Entailment & \xmark \\
    \end{tabular}
    }
    \caption{Selected qualitative examples from SNLI and MultiNLI for each split. In some cases, easy instances for SNLI are mislabeled neutral pairs while for MultiNLI easy instances are simple negations and paraphrasing. The hard split contains sentence pairs with more subtle linguistic cues.}
    \label{tab:qualitative-examples}
\end{table*}

\begin{figure*}[!hbt]
    \centering
    \includegraphics[width=0.45\textwidth]{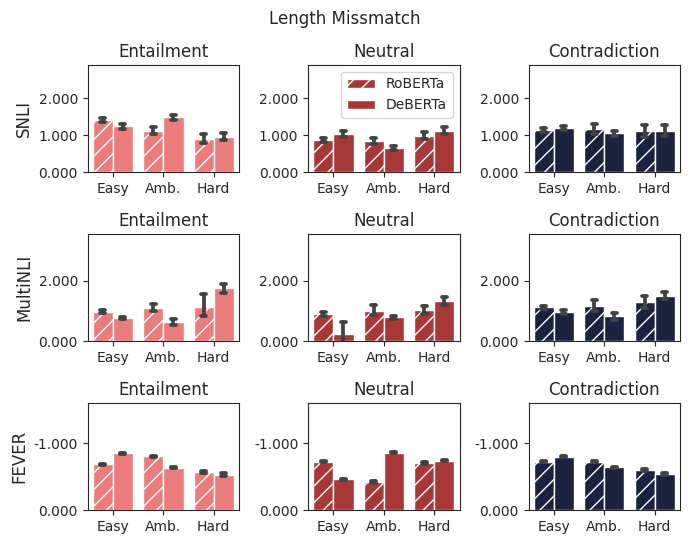}
    \includegraphics[width=0.45\textwidth]{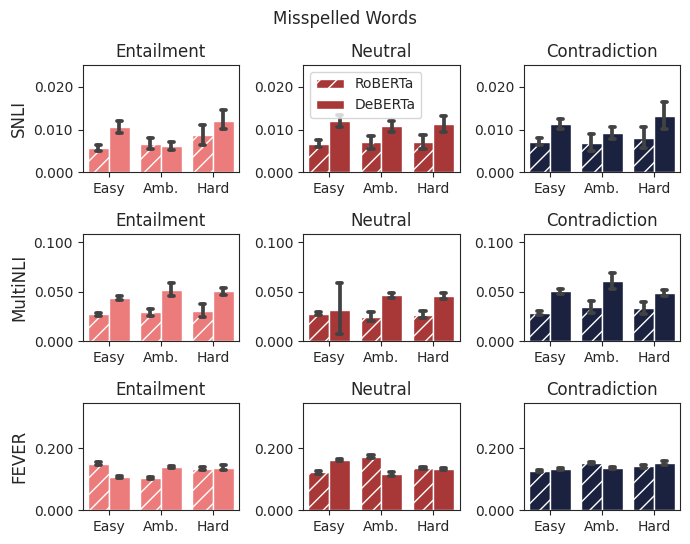}
    \includegraphics[width=0.45\textwidth]{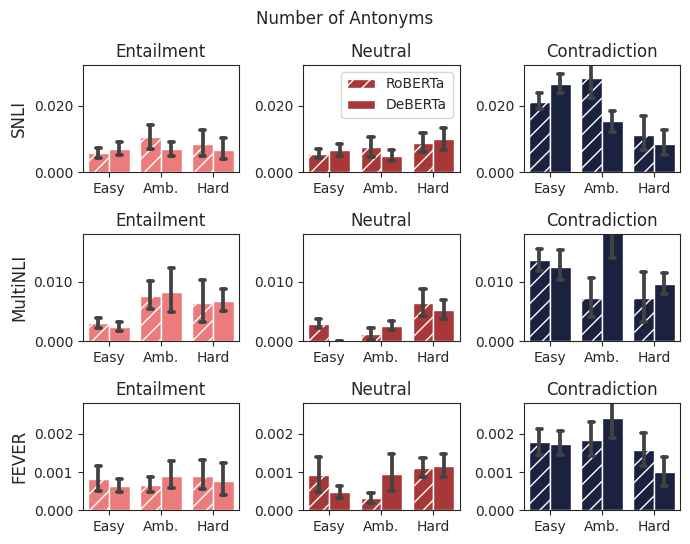}
    \includegraphics[width=0.45\textwidth]{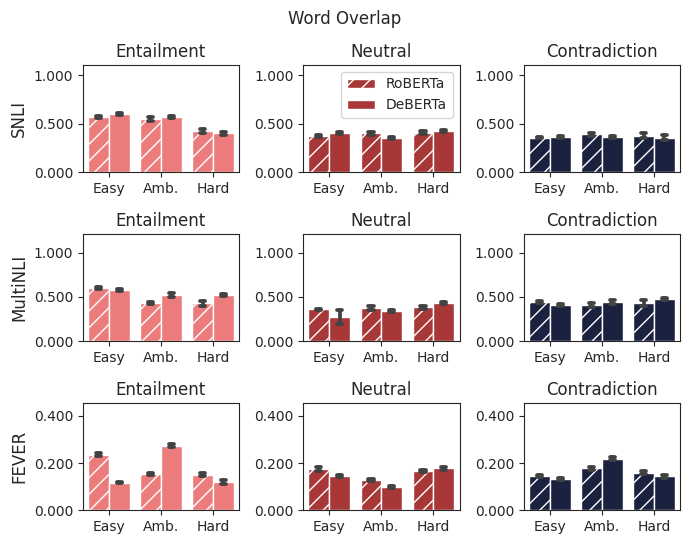}
    \caption{Extended comparison between characterizations given by RoBERTa and DeBERTa across the three datasets for the proposed heuristics.}
    \label{fig:model-comparison-extended}
\end{figure*}

\subsection{Qualitative Samples}
In Table \ref{tab:qualitative-examples}, we show qualitative examples from each of the three datasets we experimented on.

\subsection{Extended Comparison}

Figure \ref{fig:model-comparison-extended} depicts extended comparisons between the characterizations of the two models on SNLI, MultiNLI, and FEVER.

\end{document}